%% file: root.tex
\documentclass[conference]{IEEEconf}
\IEEEoverridecommandlockouts    

\usepackage{url}       
\usepackage{hyperref}
\usepackage{siunitx}
\usepackage{graphicx}
\usepackage{censor}
\usepackage{booktabs}
\usepackage{amsmath}
\usepackage{anyfontsize}
\usepackage[acronym]{glossaries}
\usepackage{placeins}
\usepackage[compress]{cite}
\usepackage{bm}
\usepackage{dblfloatfix}

\renewcommand{\vec}[1]{\bm{#1}}

\usepackage{tikz}

\newcommand\copyrighttext{%
  \footnotesize \textcopyright 2026 IEEE. Personal use of this material is permitted.
  Permission from IEEE must be obtained for all other uses, in any current or future
  media, including reprinting/republishing this material for advertising or promotional
  purposes, creating new collective works, for resale or redistribution to servers or
  lists, or reuse of any copyrighted component of this work in other works.
  DOI: \href{https://ieeexplore.ieee.org/document/11623966}{10.1109/IV66570.2026.11623966}}
\newcommand\copyrightnotice{%
\begin{tikzpicture}[remember picture,overlay]
\node[anchor=south,yshift=10pt] at (current page.south)
  {\fbox{\parbox{\dimexpr\textwidth-\fboxsep-\fboxrule\relax}{\copyrighttext}}};
\end{tikzpicture}%
\vspace{-\baselineskip}%
}

\makeglossaries
\input{acronyms.tex}

\title{\LARGE \bf Pushing the Performance Limits in Autonomous Racing: \\
Continuous Stability-Aware Adaptive Velocity Planning \\
in Formula Student Driverless}

\author{Tamara Bergerhoff, Sebastian Baader, Pascal Meißner, Frank Deinzer 
\thanks{*This work was supported by Mainfranken Racing e.V.}
\thanks{All authors are with the Center for Artificial Intelligence and Robotics (CAIRO); TUAS Würzburg-Schweinfurt, Germany \protect\\ Corresponding author: {\tt\small tamara.bergerhoff@thws.de}}%
} 

\begin{document}
	
	\maketitle
	\thispagestyle{empty}
	\pagestyle{empty}

	\input{abstract}
	\copyrightnotice
	\input{1_Introduction}

	\input{2_Related_Work}
	\input{3_Adaptive_Velocity_Planning}
	\input{4_Experiments}

	\input{5_Conclusion}

	\section*{Acknowledgments} 
	We would like to express our gratitude to the Formula Student community for its outstanding spirit of collaboration. At FSAE Italy 2025, what began with a major setback during the first discipline was turned into success thanks to tireless work within our team and the invaluable support of Schanzer Racing Ingolstadt and Hamburg Hawks. Within 24 hours, the car was repaired and returned to competition, ultimately achieving first place in DV Autocross, DV Trackdrive, DV Overall, and Overall CV-DV $|$ EV-DV. This experience was a powerful reminder of what can be accomplished through determination, teamwork, and solidarity across teams. \\ From heartbreak to triumph - this is Formula Student. $\heartsuit$

	\bibliographystyle{IEEEtran}
	\bibliography{literatur}
	
\end{document}

%% file: acronyms.tex
\newacronym{mpc}{MPC}{Model Predictive Control}
\newacronym{nmpc}{NMPC}{Nonlinear Model Predictive Control}
\newacronym{trfc}{TRFC}{Tire-Road Friction Coefficient}
\newacronym{pid}{PID}{Proportional-Integral-Derivative}
\newacronym{sgpr}{SGPR}{Sparse Gaussian Process Regression}
\newacronym{gpr}{GPR}{Gaussian Process Regression}
\newacronym{fs}{FS}{Formula Student}
\newacronym{vp}{VP}{Velocity Planning}

%% file: abstract.tex
\begin{abstract}
In autonomous racing, especially in competitions such as \textit{Formula Student Driverless}, precise planning of the target velocity of a race car is crucial for competitive lap times and stable driving behavior. 
Especially at high speeds, \gls{vp} is a significant challenge as it has to be performed in real time, taking into account track layouts, environmental influences, mechanical tolerances, and the resulting control inaccuracies.
In this paper, we present a novel approach to \gls{vp} that dynamically adapts to such changing conditions.
Instead of estimating the physical \gls{trfc}, a continuous scaling factor is inferred indirectly from vehicle stability. This factor not only reflects the effective tire-road interaction but also captures effects of control inaccuracies. From this, we generate a continuous friction map, which serves as a robust, adaptive basis for computing the optimal target speed, accounting for both vehicle and environmental limits.
Our proposed approach was evaluated on a real Formula Student race car, showing a lap time improvement of \SI{35}{\percent} over ten laps and an average increase of \SI{8}{\percent} compared to a non-adaptive approach.
\end{abstract}

%% file: 1_Introduction.tex
\glsresetall
\section{Introduction} 
Adaptive \gls{vp} is a fundamental challenge in autonomous racing, where vehicles operate at the dynamic handling limits under rapidly changing conditions. In contrast to conventional road traffic, racing environments are dominated by complex track layouts with sharp corners, small track widths, and high-speed straights. The constant pressure to achieve minimal lap times forces vehicles to operate at their physical limits. To achieve this, \gls{vp} must continuously estimate and adapt to traction limits, similar to a human driver~\cite{driverless_vs_racers}. These limits are influenced by numerous highly variable factors such as rain, temperature, and surface type, making robust adaptation challenging.

Striking the right balance between vehicle speed and stability in \gls{vp} is critical, as it serves as the key bottleneck determining overall system performance. Overly conservative strategies inevitably reduce performance, whereas overly aggressive approaches increase the likelihood of instability and loss of control. Consequently, continuous and adaptive decision-making is essential to maximize performance in dynamic environments while keeping the vehicle stable at its physical limits.

\begin{figure}[htbp] 
\centering
\includegraphics[width=0.49\textwidth]{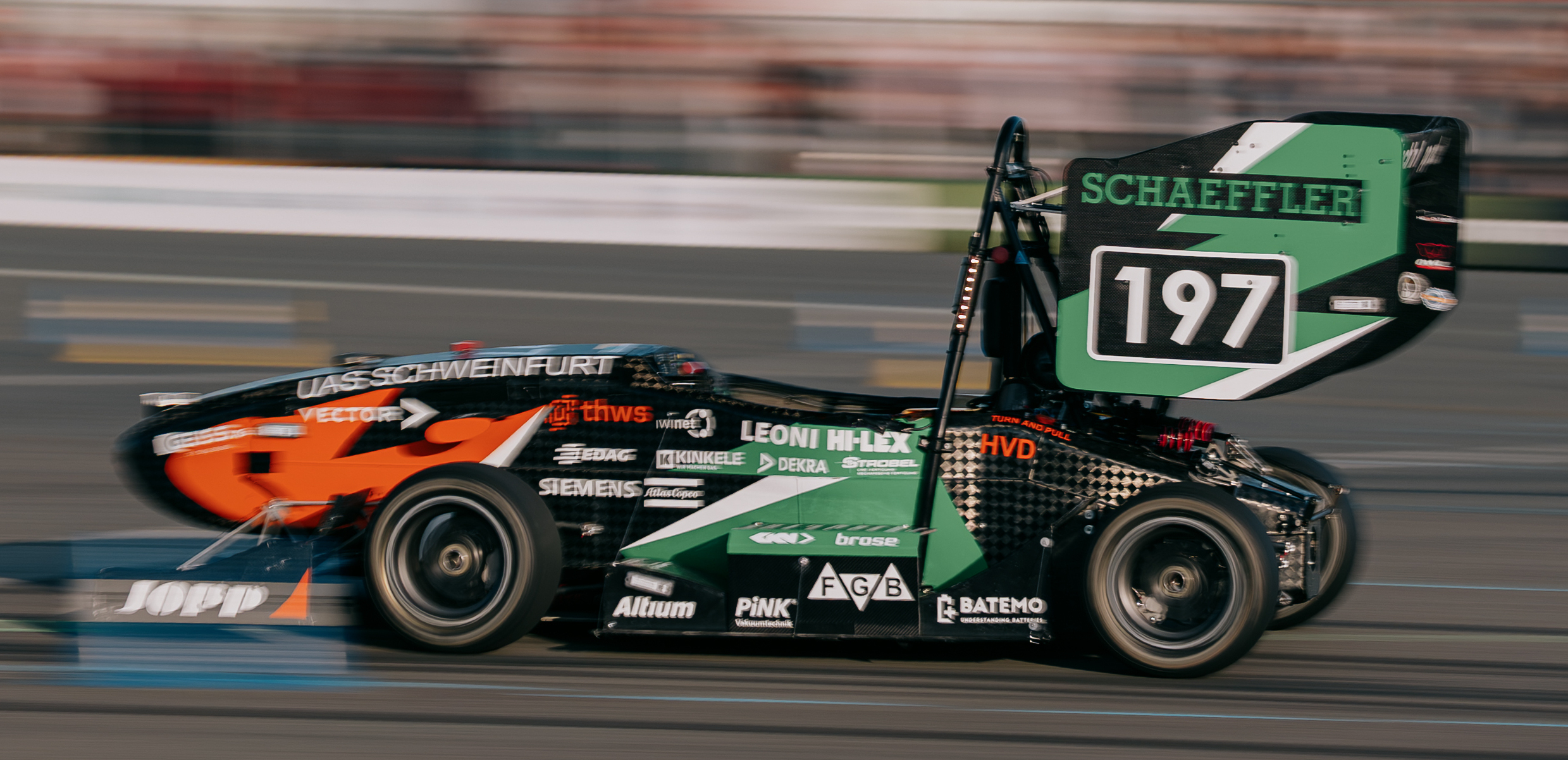}
\caption{\textit{MF17} during autonomous operation at \gls{fs} Germany 2025. \textit{MF17} is an electric, rear-wheel-drive race car capable of accelerating from 0 to \SI{100}{\kilo\meter\per\hour} in \SI{3.5}{\second}, with a top speed of \SI{137}{\kilo\meter\per\hour}. \mbox{\copyright~\gls{fs} Germany -- Photo by Maru}}
\label{fig:mf17}
\end{figure}

These challenges are evident in controlled racing environments, such as \gls{fs}, Roborace and Formula~1, where precise and continuous knowledge of grip is essential for maximizing performance. \gls{fs} provides an excellent testbed for development and evaluation, with vehicles similar to the race car shown in Fig.~\ref{fig:mf17}. 

Its circuits in \gls{fs} Driverless~\cite{FSGRules} feature straights, chicanes, and tight corners, demanding high agility and precise planning.
For example, the \textit{Trackdrive} discipline consists of 10 consecutive laps with only one vehicle on the track at a time and assesses the car's ability to operate consistently at the limits of its performance~\cite{FSGRules}. This discipline also provides an ideal opportunity to apply Adaptive \gls{vp}, as it emphasizes consistent high-speed operation under varying conditions.

In this work, we present a novel approach to \gls{vp} for autonomous race cars that dynamically adapts to changing conditions. 
The key challenge lies in computing an optimal velocity profile that defines the vehicle's velocity as a function of time or distance along a given path.
The proposed method optimizes the velocity profile by continuously incorporating historical lap information, which enables iterative refinement of the target velocity across successive laps, resulting in increasingly efficient and stable performance. Instead of explicitly estimating the physical \gls{trfc}, a continuous scaling factor is inferred indirectly from vehicle stability metrics. This factor not only reflects the effective tire-road interaction but also accounts for control inaccuracies, preventing the vehicle from attempting speeds beyond what the control system can reliably handle. From this, a continuous friction map is generated, representing the inferred scaling factor at each position along the track. 
The map effectively captures the vehicle's performance limits without explicitly estimating the physical \gls{trfc}, serving as the basis for computing the optimal target speed. 
This allows the method to seamlessly adapt to the vehicle and its control system, maximizing performance to its full potential.

%% file: 2_Related_Work.tex
\section{Related Work} 
In the context of autonomous racing, related work on \gls{vp} highlights its key role in balancing performance, stability, and safety on the race track. It has received attention in autonomous racing communities, like Roborace, the Indy Autonomous Challenge, and \gls{fs} Driverless, where high-speed maneuvers demand precise planning, as highlighted by Betz \textit{et al.}~\cite{TUMSurveyPaper}.

A common strategy is integrating \gls{vp} into \gls{mpc}, referred to as \textit{coupled control}, where the target path and velocity are planned simultaneously through steering and throttle commands~\cite{TUMSurveyPaper}. This ensures coordinated motion and avoids conflicts between path and \gls{vp}. Liniger \textit{et al.}~\cite{mini_car_mpc} demonstrated this with a nonlinear \gls{mpc} on a 1:43 scale RC race car. 
Full-scale extensions have been developed by \gls{fs} teams from Zurich and Lisbon, using vehicle models and tire constraints for more accurate dynamics~\cite{AMZFullAutonomousRacingSystem, MPCLisboa}.
Laurense \textit{et al.}~\cite{PathTrackingAtTheLimitOfFriction} propose a sideslip controller at the friction limits, which is robust under extreme tire loads but less precise during normal driving.
However, coupled approaches face the fundamental challenge of balancing precise path-tracking and lap time minimization. This complicates error parameterization, as the impact of a path deviation varies with the vehicle's speed and control inputs. Failing to account for this can compromise both stability and safety.

In contrast, \textit{decoupled control} separates steering and \gls{vp}, allowing independent handling. A simple example is a quasi-steady-state or Forward-Backward Solver~\cite{math_paper_forward_backward_solver}. Heilmeier \textit{et al.}~\cite{TUM_Forward_Backward_Solver} applied it in Roborace using a predefined GG-diagram. Similar methods are used by several \gls{fs} teams~\cite{starkstrom2022fsg, chalmersFSG, fred_paper, EForceBA}.
It is widely used due to its simplicity, computational efficiency, and robustness, making it well-suited for practical real-world applications. However, a fixed GG-diagram prevents adaptation to environmental changes, requiring conservative safety margins, which severely limit performance.

\textit{Adaptive methods} extend \textit{coupled} or \textit{decoupled control} by incorporating track-dependent friction.
Herrmann \textit{et al.}~\cite{AdaptiveVelocityOptimizationTUM} propose a \gls{nmpc}-based Adaptive Velocity Planner using high-resolution \gls{trfc} maps~\cite{ConceptForTRFCEstimation}. Christ \textit{et al.}~\cite{VariableTireRoadFrictionTUM} present a related optimal control formulation. In addition, Werner \textit{et al.}~\cite{grip_map_tum} propose a spatially resolved constraint framework for trajectory planning that accounts for local grip variations. While differing in their specific formulations, these methods generally rely on the prior availability of accurate \gls{trfc} maps. 
On permanent racetracks, such maps can often be obtained offline through repeated use, as the racing line accumulates rubber and offers higher grip, whereas track edges tend to be less favorable due to debris. In \gls{fs}, however, tracks are temporarily constructed with traffic cones, typically on unknown asphalt, and their limited width~\cite{FSGRules} makes high-resolution \gls{trfc} maps infeasible, leaving little to no prior knowledge of surface friction in this context.

When accurate \gls{trfc} maps are unavailable, \textit{adaptive methods} must rely on online or iterative estimation of the \gls{trfc}. Kapania \textit{et al.}~\cite{nitin_phd} propose a lap-time optimal planner trained under varying grip assumptions, but this requires prior track knowledge and accurate initial grip estimates. Wischnewski \textit{et al.}~\cite{ModelFreeAlgo} developed an algorithm that gradually approaches vehicle handling limits by iteratively scaling maximum feasible accelerations based on stability metrics rather than explicit vehicle models. The method applies a single scaling factor for the entire race track, effectively adjusting the GG-diagram used in \gls{vp}. Inspired by this, the \gls{fs} Team from Karlsruhe proposed an Adaptive Velocity Planner estimating the scaling factor per track section \cite{KARaceIngDavidFischer}. If the scaling factor is incorrectly set due to an erroneous stability estimate, it can significantly degrade lap time both for individual track sections and for the track as a whole.

To overcome these limitations, this paper combines the simplicity of the Forward-Backward Solver with an online estimation of the scaling factor. This yields a continuous friction map for the entire track, in which a scaling factor is stored at each position based on vehicle stability criteria, enabling the GG-diagram to adapt seamlessly to the vehicle's position and prevailing conditions.

%% file: 3_Adaptive_Velocity_Planning.tex
\section{Adaptive Velocity Planning}
\begin{figure}[h]
    \centering
    \includegraphics[width=\columnwidth]{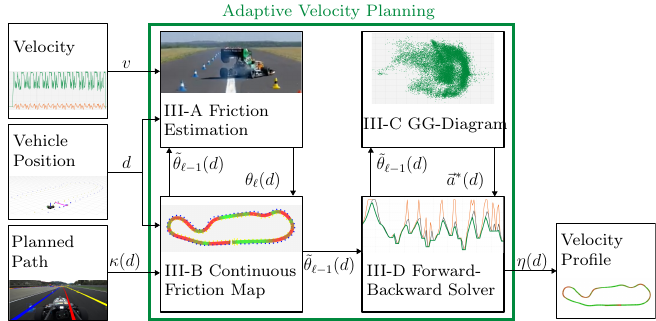}
    \caption{Overview of our Adaptive \gls{vp}. The system takes the vehicle velocity, position and the curvature of the planned path as inputs, estimates a scaling factor $\theta_\ell(d)$ based on vehicle stability metrics (\ref{subsec:friction_estimation}), updates a continuous friction map (\ref{subsec:friction_map}), and computes the acceleration limits based on $\tilde{\theta}_{\ell-1}(d)$ (\ref{subsec:gg_diagram}), which are determined for each position $d \in [d,\Lambda]$ along the prediction horizon. These limits are used by a Forward-Backward Solver to compute the velocity profile $\eta(d)$ (\ref{subsec:forward_backward}).}
    \label{fig:overview}
\end{figure}

Our proposed Adaptive \gls{vp} is illustrated in Fig.~\ref{fig:overview}. The inputs include the vehicle's current longitudinal velocity $v$ and its track progress i.e. position $d \in [0,D]$, with $D$ denoting the total lap length. Furthermore the planned track curvature $\kappa(d)$ over the upcoming track segment to be driven (approx. the next \SI{15}{\meter}), is specified. This segment is defined by the prediction horizon $[d,\Lambda]$, where $\Lambda \in [0,D]$ indicates the endpoint. The output is a velocity profile $\eta(d)$, representing the maximum feasible velocity at each point $d \in [d,\Lambda]$ along the prediction horizon.

\subsection{Friction Estimation}
\label{subsec:friction_estimation}
The first step is to estimate a scaling factor $\theta_\ell(d)$. This factor captures the local track conditions, representing how the \gls{trfc}, vehicle behavior, and control constraints influence feasible velocities. Unlike existing work, where the scaling factor is estimated either for the entire track~\cite{ModelFreeAlgo} or for track segments~\cite{KARaceIngDavidFischer}, we continuously estimate $\theta_\ell(d)$ for each position $d$.

To quantify these local conditions, a set of normalized stability values $g_j\big(\tilde{\theta}_{\ell-1}(d)\big)$ is employed, with each $j$ representing a different stability metric. These metrics can capture aspects of the vehicle's dynamic behavior, e.g. front and rear slip angles or longitudinal slip. They may also include control-related measures such as path deviation, describing the difference between the desired and actual trajectory, and state errors, quantifying discrepancies between commanded and achieved vehicle states.
 
To gradually approach the vehicle's handling limits while driving, the scaling factor is updated iteratively using a \gls{pid}-controller~\cite{astrom_pid_1995}, which adjusts $\theta_\ell(d)$ based on the discrepancy between the stability values $g_j\big(\tilde{\theta}_{\ell-1}(d)\big)$ and their target thresholds. Specifically, the error used by the \gls{pid} controller is defined as
\begin{equation}
\epsilon_\ell(d) = 1 - \max_{j} g_j\big(\tilde{\theta}_{\ell-1}(d)\big) \;\text{,}
\end{equation}
which denotes the deviation of the highest stability value from the target value $1$. Stability values below $1$ indicate stable operation, $1$ corresponds to the optimal handling limit, and values above $1$ indicate instability, requiring the vehicle to slow down in the next lap to maintain stable operation.

This ensures a smooth and controlled convergence towards the optimal friction representation for each point $d$ on the path
\begin{equation}
\theta_\ell(d) = \tilde{\theta}_{\ell-1}(d) 
+ k_P \, \epsilon_{\ell}(d) 
+ k_I \sum_{l=1}^{\ell} \epsilon_l(d) 
+ k_D \, \Delta \epsilon_\ell(d) \;\text{,}
\end{equation}
where $k_P$, $k_I$, $k_D$ are the proportional, integral, and derivative gains of the controller. This update is applied iteratively over multiple laps, tuning the scaling factor to reach a value that keeps the vehicle close to its handling limits at each track position $d \in [0,D]$.

\subsection{Continuous Friction Map}
\label{subsec:friction_map}

Scaling factors $\theta_\ell(d)$ can only be estimated for a discrete set of points $\vec{z}_\ell = [d_1,\dots,d_n]$ along the track per lap $\ell$. However, for all positions $d \in [d,\Lambda]$ within the prediction horizon a continuous map is required to obtain these values from a historic lap. Therefore, we interpolate these points using \gls{sgpr}~\cite{SGPR_original, SparseGaussian_InducingPoints}, a computationally more efficient variant of \gls{gpr} ~\cite{GaussianProcessRegressionPrimaerLiteratur}. The \gls{sgpr} relies on a set of equidistantly distributed inducing points $\vec{u} = [d_1,\dots,d_m]$ along the track. After the completion of each lap $\ell$, the measurements are used to update a sparse representation of the training data on these inducing points. To ensure continuity, the system continues to use the sparse representation from the previous lap until the new representation has been computed asynchronously, since it can only be generated after collecting all data from the current lap.

\begin{figure}[htbp]
  \centering
  \includegraphics[width=\columnwidth]{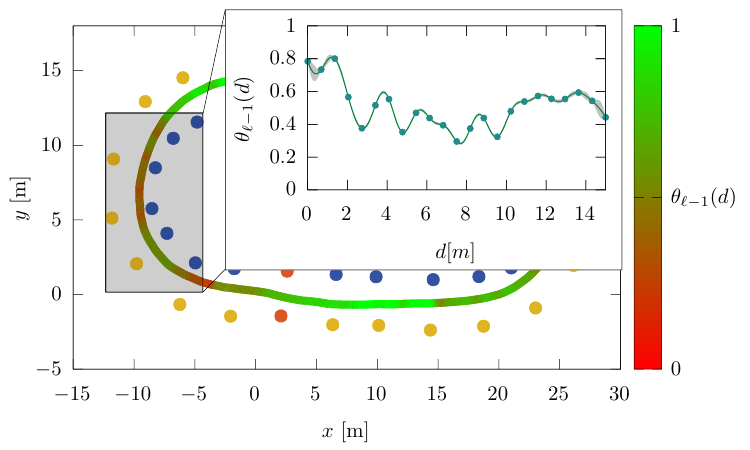}
  \caption{Visualization of a continuous friction map along one track, with color encoding the measured scaling factors $\theta_{\ell-1}(d)$. In the highlighted region we show how the \gls{sgpr} is used to retrieve interpolated values for each desired position $d \in [d,\Lambda]$ within the prediction horizon, providing smooth, continuous data for downstream \gls{vp}.}
  \label{fig:friction_map}
\end{figure}

Thus, for all positions $d \in [d,\Lambda]$ within the prediction horizon, continuous estimates $\tilde{\theta}_{\ell-1}(d)$ are obtained by
\begin{equation}
\tilde{\theta}_{\ell-1}(d) = \mathrm{SGPR}\left(d,\{\left(d,\theta_{\ell-1}(d)\right)| d \in \vec{z}_{\ell-1}\}, \vec{u}\right) \;\text{.}
\end{equation}
Fig.~\ref{fig:friction_map} illustrates this process, showing how information from a previous laps is retrieved for each position $d$ within the prediction horizon along the track. 
Our \gls{sgpr} approach allows for smooth interpolation between discrete measurement points while keeping computational costs low, which is critical for real-time \gls{vp}. This guarantees that the friction map remains available at all times, while still being refined iteratively on a lap-by-lap basis.

In contrast, existing methods typically generate continuous maps offline~\cite{AdaptiveVelocityOptimizationTUM, VariableTireRoadFrictionTUM, grip_map_tum}, rather than estimating them online during driving.

\subsection{GG-Diagram}
\label{subsec:gg_diagram}
Based on the estimated friction map, the optimal accelerations $\vec{a}^*(d) = [a_x^*(d), a_y^*(d)]^\top$ can be determined at every point $d$ along the track. These limits define the GG-diagram, which encodes the vehicle's maximum combined longitudinal $a^*_x(d)$ and lateral $a^*_y(d)$ acceleration. It illustrates the interaction between longitudinal and lateral tire forces, showing that increased utilization of friction for acceleration or braking reduces the available lateral force for cornering, and vice versa.

Depending on the chosen representation, the GG-diagram can either be scaled uniformly using a single factor $\theta_\ell(d)$, affecting both longitudinal and lateral stability, or separately using individual factors for longitudinal and lateral friction~\cite{RaceCarVehicleDynamics}. In practice, the elliptical formulation with separate scaling is mostly adopted, as it allows for a more accurate representation of the distinct longitudinal and lateral stability characteristics.

To account for estimated friction and aerodynamic effects, the acceleration limits are dynamically scaled
\begin{equation}
\vec{a}^*(d) =
\begin{bmatrix}
 \tilde{\theta}_{\ell-1}(d) \left(g + \frac{1}{m} F_\text{down}\right) - \frac{1}{m} F_\text{drag}\\
 \tilde{\theta}_{\ell-1}(d) \left(g - \frac{1}{m} F_\text{down}\right)
\end{bmatrix} \;\text{,}
\end{equation}
where $F_\text{down}$ and $F_\text{drag}$ are aerodynamic downforce and drag~\cite{RaceCarVehicleDynamics}, $m$ the vehicle mass, and $g$ the gravitational constant. In previous work, scaling factors were often estimated for the entire track~\cite{TUM_Forward_Backward_Solver, ModelFreeAlgo} or for track segments~\cite{KARaceIngDavidFischer}, which cannot capture fine-grained variations. In contrast, our approach allows the scaling factor to be continuously estimated along the path, so the GG-diagram reflects local differences in longitudinal and lateral limits.





\subsection{Forward-Backward Solver}
\label{subsec:forward_backward}
The final output, the maximum feasible velocity $\eta(d)$ along the track is computed using a Forward-Backward Solver~\cite{TUM_Forward_Backward_Solver}. 
In contrast to static formulations in existing work~\cite{TUM_Forward_Backward_Solver, ModelFreeAlgo}, the limits on longitudinal and lateral accelerations are not constant, 
but vary continuously with the position $d$ along the track according to the locally estimated GG-diagram (\ref{subsec:gg_diagram}). However, the Forward-Backward Solver can only be performed for a finite number of discrete points along the path. Therefore, the prediction horizon is discretized with a spacing of $\lambda$, and the target velocity is only computed for each of these points. This means that for every discretization point, the admissible accelerations 
$a^*_x(d)$ and $a^*_y(d)$ are different, reflecting changes in local track conditions.  

The solver operates as follows: first, curvature-dependent velocity limits $v(d)$ are obtained from
\begin{equation}
v(d) = \sqrt{\frac{a^*_{y}(d)}{\kappa(d)}} \;\text{.}
\end{equation}
The available longitudinal acceleration at each point is then derived from the local GG-diagram (\ref{subsec:gg_diagram}). 
Including the curvature-dependent lateral acceleration \mbox{$a_y(d) = (v(d))^2 \kappa(d)$}, the admissible longitudinal acceleration is
\begin{equation}
a_x(d) = a^*_{x}(d)\, \sqrt{1 - \left(\tfrac{a_y(d)}{a^*_{y}(d)}\right)^2} \;\text{,}
\end{equation}
ensuring that both longitudinal and lateral dynamic limits are consistently met. 
A forward pass then propagates feasible velocities along the track based on the local acceleration limits, while a backward pass ensures braking feasibility by enforcing deceleration limits derived from the local GG-diagram. This is performed using
\begin{equation}
\eta(d \pm \lambda) = \sqrt{(v(d))^2 + 2 a_{x}(d)\,\lambda } \;\text{.}
\end{equation}
As a result, the Forward-Backward Solver produces a velocity profile $\eta(d)$ that fully adapts to the spatially varying 
GG-diagram, ensuring that changes in track friction, aerodynamics, and vehicle stability are captured 
at every point along the track, thus enabling the vehicle to achieve higher overall speeds.

%% file: 4_Experiments.tex
\section{Experiments}
We evaluate our Adaptive \gls{vp} on the race car \textit{MF17}, shown in Fig.~\ref{fig:mf17}, in real-world experiments, comparing it to a non-adaptive baseline~\cite{TUM_Forward_Backward_Solver} and a human driver. Additionally, we benchmark it in simulation against existing methods. Finally, we present and discuss the results from the FSATA competition~\cite{FSATA_Results}.

\subsection{Adaptive vs. Non-Adaptive Comparison}
\label{subsec:adaptive_vs_non_adaptive}
We evaluated our Adaptive \gls{vp} over ten laps, comparing it to a non-adaptive \gls{vp}~\cite{TUM_Forward_Backward_Solver}. Both were validated in real-world experiments. 
\begin{figure}[htbp]
  \centering
  \includegraphics[width=\columnwidth]{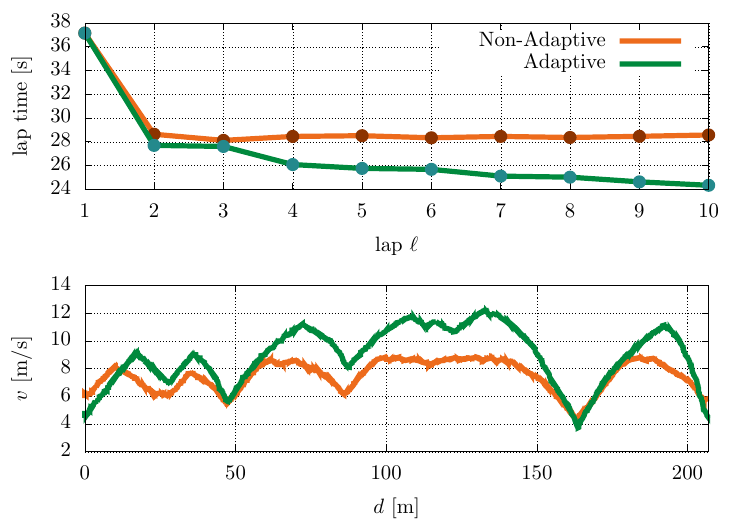}
  \caption{Comparison of our approach (green) with a non-adaptive \gls{vp} (orange). The top plot compares lap times of the two approaches over $10$ \textit{Trackdrive} laps. The bottom plot compares the velocity profiles of both approaches for the 10th lap.}
  \label{fig:adaptive_vs_non_adaptive}
\end{figure}

The top plot in Fig.~\ref{fig:adaptive_vs_non_adaptive} shows lap time progression. The initial dip in lap time after the first lap is caused by starting from a standstill. Lap times for the non-adaptive approach remain nearly constant thereafter, while our adaptive method improves continuously. It achieves a final lap time of \SI{24.32}{\second} compared to \SI{37.15}{\second} in the 1st lap, a reduction of \SI{35}{\percent}. The average lap time over $10$ laps decreases from \SI{29.30}{\second} for the non-adaptive \gls{vp} to \SI{26.90}{\second} for the Adaptive \gls{vp}, corresponding to an \SI{8}{\percent} improvement. While the non-adaptive approach operates with a fixed, conservatively chosen scaling factor that limits performance, our adaptive method continuously updates this factor for each position. This allows the vehicle to progressively exploit higher available grip, leading to continuous lap time improvements.

The bottom plot in Fig.~\ref{fig:adaptive_vs_non_adaptive} compares the velocity profiles of the 10th lap. Our Adaptive \gls{vp} clearly increases longitudinal speeds on straights while maintaining lateral accelerations in corners close to the track limits. The most notable improvements occur during acceleration and braking phases, indicating that the system successfully adapts to varying traction conditions and optimally exploits the available grip. Cornering speeds remain largely unchanged, reflecting that the vehicle is already operating near lateral limits. These results highlight that, compared to the non-adaptive approach, top speeds are improved as expected. A current limitation is that the vehicle cannot fully exploit the available grip from the first lap, as it needs to learn the track's scaling factors. 

\subsection{Assessment of Competing Methods}
\label{subsec:competing_methods}

\begin{figure}[htbp]
  \centering
  \includegraphics[width=\columnwidth]{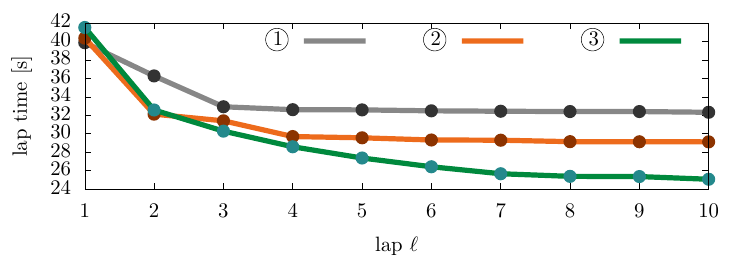}
  \caption{Comparison of lap time progression over 10 \textit{Trackdrive} laps for three approaches: Approach 1 (gray) estimates a single scaling factor for a whole lap~\cite{ModelFreeAlgo}, Approach 2 (orange) estimates one scaling factor per track segment~\cite{KARaceIngDavidFischer}, and Approach 3 (green), our method, estimates the scaling factor at every position.}
  \label{fig:lap_times_competing_methods}
\end{figure}

To further assess performance, we conducted simulation-based comparisons with established adaptive approaches presented in previous research.
We first evaluated lap times, shown in Fig.~\ref{fig:lap_times_competing_methods}.

Three approaches were compared: a single scaling factor for the entire lap~\cite{ModelFreeAlgo}, a scaling factor per track segment~\cite{KARaceIngDavidFischer}, and our continuous approach. 

\begin{figure}[htbp]
  \centering
  \includegraphics[width=\columnwidth]{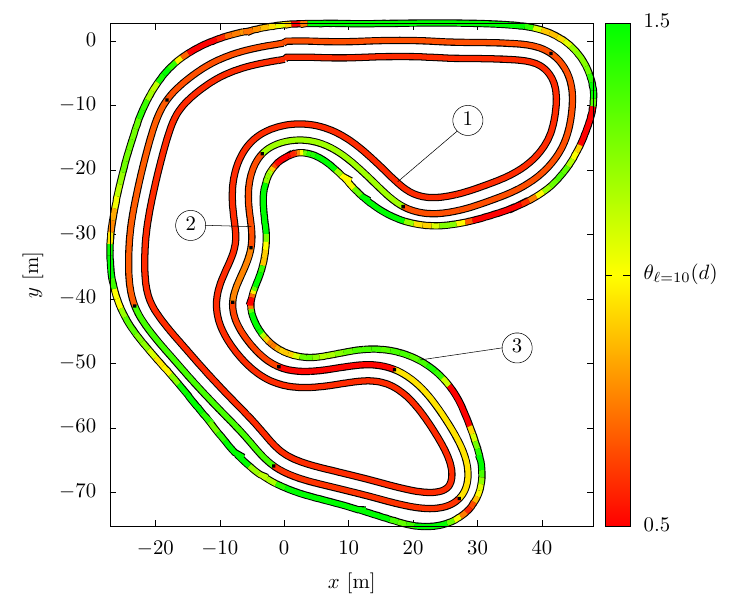}
  \caption{Comparison of estimated friction maps for three approaches: Approach 1 estimates a single scaling factor~\cite{ModelFreeAlgo}, Approach 2 estimates one scaling factor per track segment~\cite{KARaceIngDavidFischer}, and Approach 3, our method, continuously estimates the scaling factor at every position.}
  \label{fig:friction_map_competing_methods}
\end{figure}

Both the single-factor approach and the segment-based method are constrained whenever the vehicle detects instability at a specific point on the track. For instance, if unstable behavior occurs at a particular position, the scaling factor is reduced for the entire segment or lap, which in turn limits the achievable velocity at other positions along the track. This is particularly visible in Fig.~\ref{fig:friction_map_competing_methods}.

In Approach 1, improvements in the scaling factor are mainly limited by cornering, leading to a less accurate approach to the desired value, while in Approach 2, similar limitations occur due to segment-based adjustments. Both approaches can become potentially dangerous under abrupt scaling factor changes and can even lead to a crash, as noted in~\cite{crash_paper}. A sudden reduction in scaling factor at a segment or lap end can cause the vehicle to lose stability and slide out, which can occur both during braking and acceleration. In contrast, our method uses a continuous map, as illustrated in Fig.~\ref{fig:friction_map_competing_methods}, where the scaling factor varies smoothly along the track while avoiding the abrupt changes seen in Approach 2, e.g. at $(-1, -66)$[m], where the scaling factor differs by almost $1$.

This reduces the likelihood of extreme abrupt friction changes. Overall, both the lap times and the friction maps indicate that our approach provides a more realistic representation of vehicle limits, allowing for higher speeds and improved lap times. 


\begin{table*}[!t]
    \centering
    \vspace{0.2 cm}
     \caption{Lap Time Comparison at FSATA Competition~\cite{FSATA_Results}}
    \label{tab:laptimes_total}
    \begin{tabular}{l*{10}{c}c}
        \toprule
        \textbf{Team} & \multicolumn{10}{c}{\textbf{Lap Time} [s]} & \textbf{Total} [s] \\
        \cmidrule(lr){2-11}
                      & Lap 1 & Lap 2  & Lap 3  & Lap 4  & Lap 5  & Lap 6  & Lap 7  & Lap 8  & Lap 9 & Lap 10 &  \\
        \midrule
        Us           & 27.344 & 25.462 & 25.335 & 24.887 & 24.776 & 24.593 & 24.585 & 24.532 & 24.605 & 25.054 & 251.178 \\
        Competitor 1 & 40.925	& 40.854 & 41.052	& 40.975 & 40.905	& 41.105 & 41.015	& 40.786 & 40.591	& 40.702 & 408.913 \\
        Competitor 2 & 51.154	& 41.569 & 42.234 &	42.349 & 42.176 &	41.834 & 41.734 &	40.665 & 41.146	& 40.912 & 425.777 \\
        \bottomrule
    \end{tabular}
\end{table*}

\subsection{Human Driver vs. Driverless}
\label{subsec:hd_vs_dv}
To assess how close our Adaptive \gls{vp} approaches the handling limits of the vehicle, we compared the performance with a human driver. Both the driver and the driverless system completed a \textit{Trackdrive}. The driver had no prior knowledge of the track and had never driven the race car before, similar to the driverless system.
The driverless system followed the center line without optimizing the racing line, which could slightly bias the results.
\begin{figure}[htbp]
  \centering
  \includegraphics[width=\columnwidth]{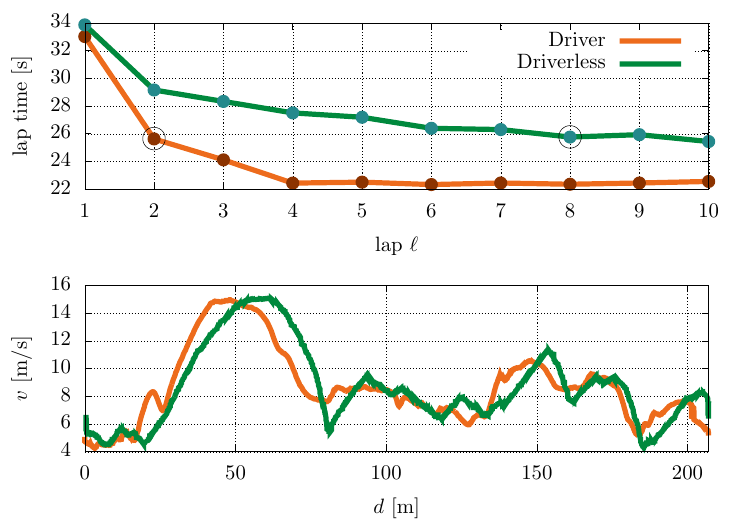}
  \caption{Comparison of our approach (green) with a human driver (orange). The top plot shows lap time progression over $10$ \textit{Trackdrive} laps. The bottom plot compares the velocity profiles for the laps with the most similar lap times. The driver's 2nd lap and our approach's 8th lap.}
  \label{fig:human_driver_vs_driverless}
\end{figure}

Fig.~\ref{fig:human_driver_vs_driverless} shows lap time progression. Both achieved similar times on the first lap, with the driver initially skidding due to underestimating track friction. The driver improved much faster, reaching his optimum lap time around lap $3$, whereas the driverless system improved gradually, closing the gap to the human driver's lap time to roughly \SI{3}{\second} by the last lap. The human driver benefits from prior knowledge and experience, allowing him to anticipate grip limits and intuitively know how to drive. He can also recover more easily when overestimating traction, which enables riskier maneuvers. This suggests that incorporating experience-based data could improve the driverless system. 

Despite this difference in learning behavior, the velocity profiles for the most comparable laps (driver's 2nd vs. our 8th) are very similar, showing that the driverless system closely replicates the driver's speed patterns, though it accelerates slightly more gradually and brakes more aggressively. Cornering and top speeds are similar, indicating that our method accurately exploits available traction.

\begin{figure}[htbp]
  \centering
  \includegraphics[width=\columnwidth]{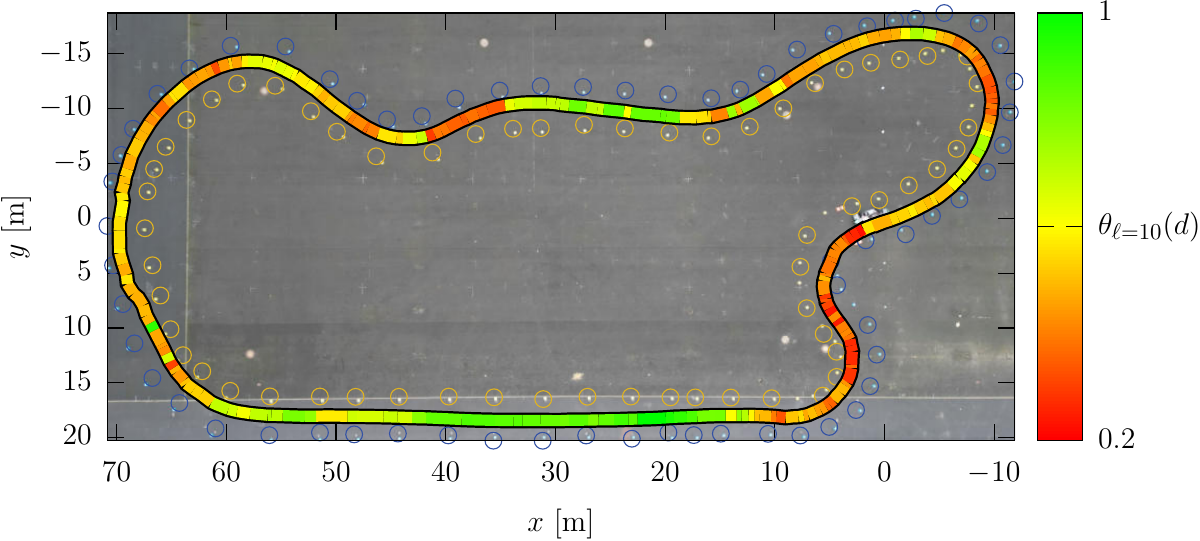}
  \caption{Friction map overlaid on a drone image of the race track. The track features three asphalt types, making driving particularly challenging.}
  \label{fig:drone_capture_friction_map}
\end{figure}

The track driven by both the driverless system and the human driver is shown in Fig.~\ref{fig:drone_capture_friction_map}. The track features multiple asphalt types, creating particularly challenging conditions. The friction map highlights reduced traction at transitions between different surfaces, with lower values extending slightly before and after the transition due to the vehicle's physical length and dynamics. The map corresponds to the 10th lap of the driverless system, showing that cornering sections consistently exhibit lower traction than straights. This suggests that the Adaptive \gls{vp} correctly anticipates and adapts to areas of limited grip, maintaining stability while maximizing speed where the track allows.

\subsection{Performance Under Competition Conditions}
\label{sec:fsata}

Finally, we also evaluated our Adaptive \gls{vp} during the FSATA competition~\cite{FSATA_Results} using the \textit{MF17}, completing $10$ laps under real-world competition conditions. Table~\ref{tab:laptimes_total} summarizes the lap times for our vehicle and two competitors.

Our approach achieved a total lap time of \SI{251.18}{\second}, outperforming Competitor 1 (\SI{408.91}{\second}) and Competitor 2 (\SI{425.78}{\second}). Our lap times improved over the first laps as the system adapted to the track and optimized the velocity profile, resulting in an improvement of \SI{10}{\percent}. The final lap was slightly slower due to accidental wheel lock-up while braking on sections with two different asphalt types, which limited deceleration and required a more cautious approach. Additionally, the maximum velocity was capped at \SI{14}{\meter\per\second} for safety reasons, meaning that only cornering, braking, and acceleration could be optimized, which naturally limited the achievable improvements on our end. 

Competitor 1, in contrast, showed no improvement in lap times at all. There was also no improvement from lap $1$ to lap $2$ because the light barrier is positioned before the start line, allowing the vehicle to reach its top speed within this distance. Competitor 2, however, exhibited a substantial difference between lap $1$ and lap $2$, as the initial lap was primarily used to acquire knowledge of the track before effectively optimizing their velocity profile. 


Overall both competitors, showed slower and more inconsistent lap times, highlighting the advantage of Adaptive \gls{vp} under varying track conditions.

%% file: 5_Conclusion.tex
\glsresetall
\section{Conclusion}
This paper addressed the challenge of real-time \gls{vp} for autonomous race cars under changing conditions. We presented an approach that builds a continuous, stability-based friction map online during driving to optimize the target velocity, accounting for both vehicle performance limits and control inaccuracies.

The experimental evaluation demonstrates that our Adaptive \gls{vp} significantly improves both lap times and velocity profiles compared to a non-adaptive baseline, while closely approaching the performance of a human driver. The system effectively increases longitudinal speeds on straights, maintains lateral accelerations near track limits in corners, and adapts to varying traction conditions, as evidenced in Sec.~\ref{subsec:adaptive_vs_non_adaptive} and \ref{subsec:hd_vs_dv}. Simulation results in Sec.~\ref{subsec:competing_methods} confirm that space-continuous estimation of the scaling factor allows higher speeds and safer handling than single-factor or segment-based methods, particularly under abrupt friction changes. Real-world experiments at the FSATA competition further validate the method (Sec.~\ref{sec:fsata}), showing consistent improvements over competitors and robust performance on a track with multiple asphalt types.

Overall, the approach is not only suitable for Formula Student but could also support manual driving by providing a stability-aware map for enhanced torque vectoring. Such high-traction maps and adaptive velocity optimization could further benefit high-performance applications, including Roborace and Formula 1, where precise knowledge of grip and track conditions is essential for maximizing vehicle performance, leaving room for future work.